\pdfoutput=1

\documentclass[11pt]{article}

\usepackage{ACL2023}
\usepackage{bbm}

\usepackage{times}
\usepackage{latexsym}
\usepackage{enumitem}
\usepackage{booktabs}
\usepackage{centernot}
\usepackage{amssymb}

\usepackage{multirow} 

\usepackage{listings}

\usepackage{algorithmic}

\usepackage[T1]{fontenc}

\usepackage[utf8]{inputenc}
\usepackage{mathtools}
\usepackage{amsmath}

\usepackage{microtype}

\usepackage{inconsolata}

\usepackage{makecell} 
\usepackage{siunitx} 

\usepackage{caption} 
\usepackage{threeparttable} 

\def\cM{\mathcal{M}}

\usepackage{etoolbox}


\newtheorem{theorem}{Theorem}[section]

\newtheorem{definition}[theorem]{Definition}
\newtheorem{assumption}[theorem]{Assumption}

\newenvironment{proof}[1][Proof]{\begin{trivlist}
\item[\hskip \labelsep {\bfseries #1}]}{\end{trivlist}}

\title{WatME: Towards Lossless Watermarking Through Lexical Redundancy}

\author{Liang Chen$^{{\spadesuit}}$~~Yatao Bian$^{\heartsuit}$~~Yang Deng$^{\diamondsuit}$~~Deng Cai$^{\heartsuit}$ \\
\textbf{Shuaiyi Li}$^{\spadesuit}$\textbf{~~Peilin Zhao}$^{\heartsuit}$\textbf{~~Kam-Fai Wong}$^{\spadesuit}$ \\
$\spadesuit$ The Chinese University of Hong Kong 
$\heartsuit$ Tencent AI Lab 
$\diamondsuit$ National University of Singapore \\ 
\texttt{\{lchen, kfwong\}@se.cuhk.hk}\\
}

\begin{document}
\maketitle

\begin{abstract}\vspace{-0.05cm}
Text watermarking has emerged as a pivotal technique for identifying machine-generated text. 
However, existing methods often rely on arbitrary vocabulary partitioning during decoding to embed watermarks, which compromises the availability of suitable tokens and significantly degrades the quality of responses.
This study assesses the impact of watermarking on different capabilities of large language models (LLMs) from a cognitive science lens. Our finding highlights a significant disparity; knowledge recall and logical reasoning are more adversely affected than language generation. These results suggest a more profound effect of watermarking on LLMs than previously understood.
To address these challenges, we introduce Watermarking with Mutual Exclusion (WatME), a novel approach leveraging linguistic prior knowledge of inherent lexical redundancy in LLM vocabularies to seamlessly integrate watermarks. Specifically, WatME dynamically optimizes token usage during the decoding process by applying a mutually exclusive rule to the identified lexical redundancies. This strategy effectively prevents the unavailability of appropriate tokens and preserves the expressive power of LLMs. 
We provide both theoretical analysis and empirical evidence showing that WatME effectively preserves the diverse capabilities of LLMs while ensuring watermark detectability.
Our code will be released to facilitate future research.
via \url{https://github.com/ChanLiang/WatME}.
\end{abstract}

\section{Introduction}
The advent of large language models~\cite{Ouyang2022TrainingLM,OpenAI2023GPT4} with human-level generative capabilities presents tremendous opportunities across diverse domains \citep{deng-etal-2023-prompting,li2024consecutive,wang2023survey}. However, their ability to synthesize high-quality text also raises widespread concerns about potential misuse, including the dissemination of misinformation ~\cite{zellers2019defending,chen-etal-2023-beyond} and the facilitation of academic dishonesty ~\cite{StokelWalker2022AIBC}. This necessitates developing techniques to reliably attribute generated text to AI systems.

\begin{figure}[t]
    \centering
    \includegraphics[width=\linewidth]{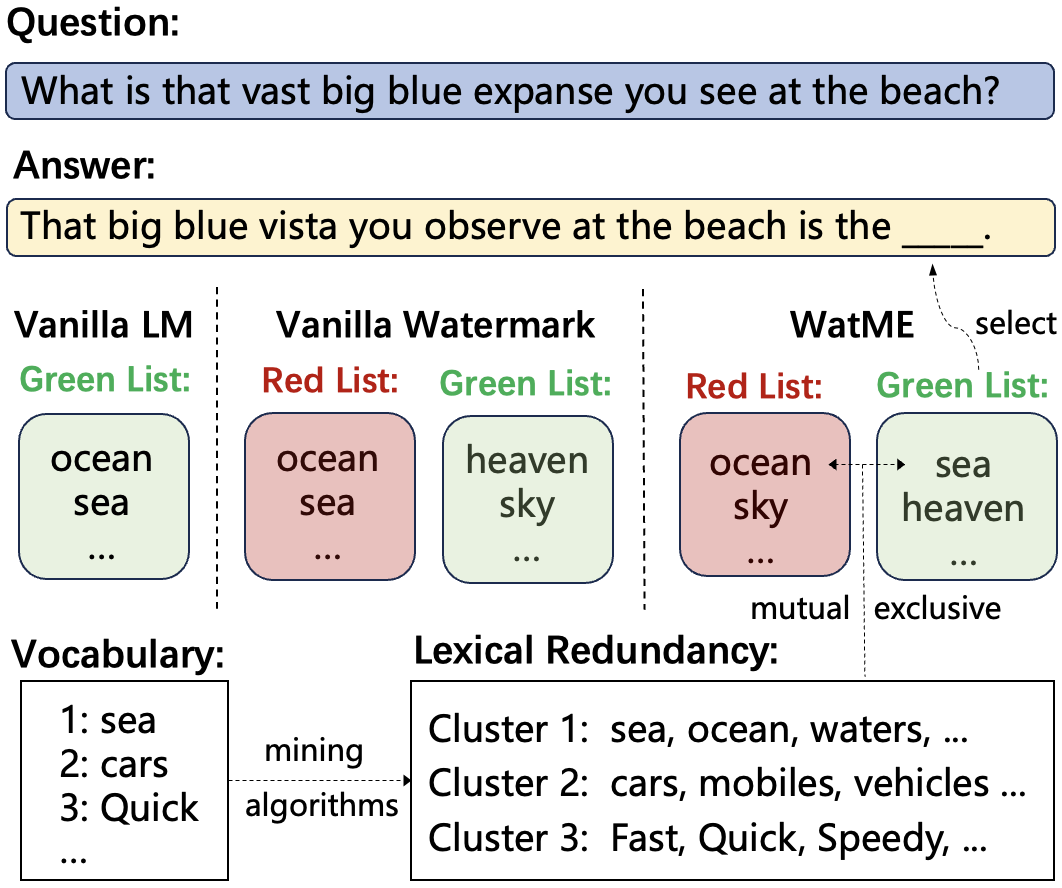} 
    \vspace{-0.60cm}
    \caption{An illustration of WatME's advantage for lossless watermarking. The left panel depicts a vanilla LM with all words available during generation. The middle panel exposes the flaw in vanilla watermarking, which may assign all suitable tokens (e.g., 'ocean' and 'sea') to the red list, diminishing text quality. The right panel underlines how WatME exploits lexical redundancy by applying a mutual exclusion rule between such words, ensuring at least one suitable word remains on the green list, thereby improving text quality.}
    \label{fig:framework}
  \vspace{-0.30cm}
\end{figure}

Existing approaches typically fall into two main paradigms. The first type attempts to distinguish machine-generated text by hunting for inductive statistical or linguistic patterns \cite{Gehrmann2019GLTRSD,Mitchell2023DetectGPTZM,zellers2019defending,OpenAI2023Detect}, employing methods that span from basic manual feature engineering to the intricate training of complex classifiers. 
However, as generative models continue improving, their outputs increasingly resemble the pattern of human writing, rendering statistical detectors ineffective \cite{dou2022gpt3, Sadasivan2023CanAT,chakraborty2023possibilities}. The second paradigm promotes a more proactive approach, advocating for direct intervention in the generative process to actively watermark model outputs \cite{watermark, christ2023undetectable,zhao2023provable}. This strategy embeds identifiable fingerprints within machine-generated text, enabling provenance verification. As LLMs' capabilities continue to grow, this approach is more effective in detecting LLM-generated text \cite{Sadasivan2023CanAT}. However, introducing watermarks during text generation can significantly impact output quality, which has been a consistent challenge for model developers - how to effectively watermark while preserving text quality.

Recent studies have attempted to improve text quality by ensuring unbiased output distributions in watermarking \cite{stanford2023robust,hu2024unbiased}, while employing pseudorandomness-guided perturbations or reweighting to adjust the original output distributions of LLMs. However, an unbiased distribution in expectation does not guarantee high text quality, and the use of these techniques reduces the effectiveness of watermark detection, especially in models that have undergone alignment training \cite{stanford2023robust}, thereby diminishing the practical utility of these methods.

In this paper, we introduce a novel approach to text watermarking by leveraging engineered lexical redundancy during the decoding phase of language generation. Our method utilizes the comprehensive set of tokens available to a language model, clustering them based on overlapping semantic or syntactic functionalities to create sets of interchangeable tokens. This process simulates redundancy within the lexical space, akin to the surplus pixels in images that facilitate watermarking in multimodal data \cite{digital_watermark,digital_watermark2}. The motivation for this strategy arises from the challenge of applying traditional watermarking techniques to textual data. In contrast to the inherent redundancy found in images, the discrete and succinct nature of textual data offers little to no native redundancy, making it challenging to exploit redundancy in the textual space \cite{zho1u-etal-2021-defense,He_Xu_Lyu_Wu_Wang_2022}. By engineering lexical redundancy, 
our method not only surmounts the limitations imposed by the inherent properties of natural language but also paves the way for secure and efficient text watermarking.

After exploring these redundancies, we exploit them via our novel algorithm, WatME, which enhances text quality by integrating a mutual exclusivity rule within the context of lexical redundancy during the watermarking process.
Specifically, WatME refines the decoding process by explicitly assigning words within each redundant cluster to distinct 'green' or 'red' teams, ensuring that no single cluster is wholly allocated to one team.
Our approach confers two main advantages: (1) it enables the 'green' team to capture a broader array of semantics, thereby boosting the model's expressive power; and (2) it increases the probability that the LLM selects the most appropriate word at each decoding step, e.g., in Figure \ref{fig:framework}, vanilla watermarking can assign all suitable words to the 'red' list, thus severely impairing performance. In contrast, our approach guarantees the presence of at least one appropriate word, thus preserving the model's expressiveness.
Building on these methodological advances, extensive theoretical and empirical evidence supports their effectiveness without compromising detection capabilities. These improvements significantly bolster the emergent abilities of large models under watermarks, surpassing the performance of baseline methods.

Our main contributions are as follows: 
\begin{itemize}[leftmargin=*]
    \item Motivated by multimedia data's inherent redundancy and the precise conciseness of text, we propose two distinct approaches for mining \textbf{\textit{lexical redundancy}}.
    \item We develop the WatME algorithm, which embeds mutual exclusion rules within the lexical space for text watermarking. Theoretical analysis is presented to validate its effectiveness in preserving the quality of text responses.
    \item Experimental results show that WatME effectively outperforms existing methods in retaining the emergent capabilities of LLMs, notably knowledge recall and logical reasoning,  within the conceptual framework of Cattell's cognitive theory, without compromising detectability.
\end{itemize}

\section{Related Work}

Early works on AI-generated text detection develop post-hoc detection methods to analyze machine-generated text by treating the problem as a binary classification task \cite{openaidetectgpt2, Jawahar2020AutomaticDO, Mitchell2023DetectGPTZM}. For instance, OpenAI has fine-tuned RoBERTa \cite{liu2019roberta} to distinguish between human and GPT-2 generated texts \cite{openaidetectgpt2}. However, existing detectors are found to be fragile against adversarial attacks \cite{Wolff2020AttackingNT} and biased towards non-native English writers \cite{Liang2023GPTDA}. Moreover, as LLMs continue to advance, their generated outputs more closely resemble human-written text, rendering these methods progressively less effective.

On the other side,  watermarking, traditionally a copyright marking method \cite{adi2018turning, rouhani2018deepsigns}, involves developers, users, and regulatory entities. Developers choose an algorithm to subtly embed hidden modifications into data, which can be altered during user transmission. Regulatory bodies can later extract this information to trace and regulate AI-generated content \cite{atallah2001natural, wilson2014linguistic,regulate}.
In the context of natural languages, watermarking 
typically involves modifying content or structure. For example, rule-based methods \cite{Kutter2000InformationHT} or carefully designed neural encoders \cite{yang2022tracing, ueoka-etal-2021-frustratingly} encrypt messages into text, which are then extracted using the corresponding rules and neural decoder. The discrete nature of natural language, however, presents a considerable challenge to this approach, as modifications can unintentionally degrade text quality or alter its intended meaning.

For the detection of LLM-generated texts, a pioneering watermarking technique \cite{watermark} partitions tokens into 'green' and 'red' lists, biases output distribution towards 'green' tokens, and creates patterns that are detectable yet imperceptible to humans.
Nevertheless, while yielding promising detection results, these methods may still degrade the textual quality and be vulnerable to the paraphrase attack.
Current efforts \cite{christ2023undetectable,fernandez2023bricks,zhao2023provable} in this field aim to develop more robust watermarking methods capable of defending various user attacks.

Apart from improving robustness, a few studies have recognized the importance of enhancing the quality of text produced by watermarked LLMs. 
\cite{stanford2023robust} utilizes Gumbel softmax to incorporate pseudorandomness-based randomness into the output distribution of language models. While this technique alters the probability distribution, the Gumbel softmax ensures that the expected distribution remains approximately unchanged, thus rendering the watermarking process unbiased.
Recent work \cite{hu2024unbiased} also shares a similar philosophy of employing reweighting technology for unbiased output distribution transformations, preserving the expected distribution unbiased.
However, unbiased distribution can not guarantee unaffected text quality. Furthermore, these methodologies have shown a marked decrease in detection performance, particularly for aligned LLMs \cite{stanford2023robust}.
Addressing these shortcomings, our research introduces a novel paradigm that exploits the intrinsic redundancy in the text generation process of LLMs to create more lossless watermarks, with a special emphasis on LLMs' emergent capabilities, thereby offering a watermarking solution that is both lossless and consistently detectable.

\section{Method}
In this section, we begin by providing a summary of the preliminaries related to text watermarking. Subsequently, we delve into an investigation of redundancy in the lexical space and demonstrate how this redundancy can be leveraged to develop a watermarking algorithm that achieves a higher degree of losslessness for large language models. Finally, we employ mathematical analysis to elucidate the benefits of our proposed method.

\subsection{Preliminary}
The watermarking process is composed of two fundamental procedures: watermark encoding and watermark detection. The encoding procedure is carried out by developers to insert a watermark into an output natural language sequence $\boldsymbol{y}$, generated by a LLM $\mathcal{M}$ for a given prompt $\boldsymbol{x}$. While the detection procedure, performed by regulators, involves the extraction and identification of the watermark from the sequence $\boldsymbol{y}$ for the purpose of monitoring the output from model $\mathcal{M}$. The algorithms that detail these procedures are described in the Appendix \ref{sec:algorithms}.

The watermark encoding process is guided by two parameters: $\gamma$ and $\delta$. At each decoding step $t$, it uses a hash key, which could be the index of the previous token, to partition the vocabulary $\mathcal{V}$ into two subsets: a green list $G_t$ which encourages usage, and a red list $R_t$ which discourages usage. The parameter $\gamma$ determines the size of the green list, while $\delta$ specifies the degree of encouragement for the green list, the increase in current logits $\boldsymbol{\ell}_t$ before performing softmax, as Eq.\ref{eq:encoding}. As $\delta$ rises, the watermark becomes more detectable in the subsequent detection process, but it may also compromise the quality of the generation. In real-world regulatory scenarios, where high detectability is required, a large $\delta$ value is generally preferred.
\begin{equation} \label{eq:encoding}
\begin{aligned}
\hat{\boldsymbol{\ell}}_t[i] &:= \boldsymbol{\ell}_t[i] + \delta, && i \in G_t \\
\hat{\boldsymbol{p}}_t &= softmax(\hat{\boldsymbol{\ell}}_t)
\end{aligned}
\end{equation}

The watermark detection process counts the number of green list tokens within $\boldsymbol{y}$, denoted by $|\boldsymbol{y}|_G$, using Eq.\ref{eq:green_size}. This process begins with the null hypothesis \textit{$H_0$: The text sequence is generated without adherence to the green list rule}. A $z$-statistic is then computed by Eq.\ref{eq:z-score}. If the $z$-score surpasses a pre-specified threshold, the null hypothesis is rejected, and the watermark is identified. 
\begin{align}\label{eq:green_size}
|\boldsymbol{y}|_G &= \sum\nolimits_{t=1}^n \mathbbm{1}(y_t \in G_t),\\
\label{eq:z-score}
z_{\boldsymbol{y}} &= \left(|\boldsymbol{y}|_G-\gamma |\mathcal{V}|\right) / \sqrt{|\mathcal{V}| \gamma(1-\gamma)}.
\end{align}

\subsection{Explore the Redundancy in Lexical Space}

\paragraph{Concept of Lexical Redundancy} Inspired by the success of image watermarking, we hypothesize that identifying redundancy within data can enable watermarking that doesn't compromise text quality. We thus explore the same opportunities within textual data, a challenging task given the discrete nature of natural language.

To address this challenge, we introduce a related concept in NLP: \textbf{\textit{lexical redundancy}}. This phenomenon arises during text generation when the most appropriate word is selected from a large, pre-constructed vocabulary. Often, this vast vocabulary includes numerous words with similar semantic and syntactic functions — a feature that makes these words interchangeable, thereby resulting in the inherent redundancy in the lexical space. 

Our interest in exploring lexical redundancy is grounded in the understanding that interchangeable synonyms, even when used in varied contexts, can deliver similar or identical semantic or syntactic functions. This insight assists in preserving the quality of text generation through an optimized watermark encoding method. For instance, if a suitable word is allocated to the red list, while its synonym is placed in the green list, then the language model can still express the intended semantics or accomplish the necessary syntactic functions.  This understanding forms the primary motivation for investigating lexical redundancy.
 
\noindent
\paragraph{Constructing Redundant Lexical Clusters}
To this end, we now focus on the construction of lexical redundancy. This process involves automatically grouping tokens—each with similar semantic or syntactic functions—from the language model's vocabulary into clusters. Each cluster, made up of interchangeable tokens, is designed to express a specific semantic or syntactic unit. 

To obtain high-quality redundant lexical clusters, we propose the following two different methods: the dictionary-based method, and the prompting-based method:

\begin{itemize}[leftmargin=*]
   \item \textbf{Dictionary-Based Method:} Utilize external dictionaries, such as WordNet \cite{wordnet} and Youdao Dictionary, to discover synonyms within the vocabulary. These synonyms often can be substituted for each other, although there are inevitably some cases where they cannot be interchanged due to polysemy. This method is beneficial for exploiting established synonym relationships but is limited to complete words due to its dependency on external resources. 
    \item \textbf{Prompting-based Method:} We prompt large language models, such as LLaMA2 \cite{llama2}, to infer synonyms for a given token by utilizing in-context learning techniques \cite{gpt3}, with the demonstrations being annotated manually by us. Detailed prompts are deferred to  Appendix \ref{cluster_mining_prompt}.
\end{itemize}

To acquire higher-quality clusters with fully interchangeable tokens, we employed two strategies during the mining process:

\paragraph{Handling Subword Tokenization}

Subword tokenization blends word and character-based approaches \cite{bpe,wordpiece,sentencepiece}, challenges the mining of redundant lexical clusters in neural text processing. This technique typically retains common words as full units and decomposes rare words into subunits. Our research mitigates these challenges by \textbf{\textit{concentrating on intact, frequently used words during preprocessing}}, thereby diminishing noise and simplifying the algorithm. 

\paragraph{Incorporating Grammatical Factors}

In the context of English, the identification of interchangeable words demands consideration of grammatical factors—tense, voice, and number—alongside semantic similarity. For instance, 'car' and 'vehicles' differ in number, affecting interchangeability. Our method addresses these issues by implementing a rule set that screens for grammatical inconsistencies, ensuring the generation of coherent and high-quality lexical clusters for subsequent use.

These strategies yield lexical clusters, with each row in Figure \ref{fig:framework}'s bottom right panel representing a cluster of interchangeable tokens. Cluster quality is manually evaluated in Section \ref{sec:cluster_quality}.

\subsection{WatME: Exploit the Lexical Redundancy}

Having constructed redundant clusters within the lexical space, 
we now turn to exploit these for a lossless watermark algorithm.

To facilitate the description of our algorithm, we provide some definitions: A subset $S \subseteq \mathcal{V}$ is defined within the vocabulary $\mathcal{V}$ of a language model $\mathcal{M}$. This subset specifically comprises complete tokens that share synonyms within the vocabulary. We then denote a collection of redundant lexical clusters we mined as $C = \{C_i \mid i=1..n\}$ such that $\bigcup_{i=1}^{n} C_i = S$. Each cluster, $C_i$, is represented as a token collection $C_i = \{s_{ij} \mid j = 1..m_i, s_{ij} \in S\}$ for $i = 1..n$, and any pair of tokens $s_{ij}, s_{ik} \in C_i$ are interchangeable.
We propose to implement our understanding of lossless watermarks by introducing  \textbf{\textit{a mutual exclusion rule}} building on the identified lexical clusters: interchangeable tokens are mutually exclusive during partitioning. In other words, if a fraction of tokens $\mathcal{A}$, representing a certain semantic or syntactic function, is assigned to the red list, then their remaining synonyms $\mathcal{B}$ should be placed on the green list, and vice versa. 

We then detail the WatME encoding process, outlined in Alg. \ref{algo:WatME-encode}, which employs a two-step partitioning process to form a green and red list. The first partition occurs within the redundant lexical clusters $C$ that we have identified within $S$, while the second takes place among the remaining part in the vocabulary denoted as $\mathcal{V} \setminus \mathcal{S}$. 
We use $\gamma$ to determine the number of tokens from the mined clusters that are allocated to the green list $G_{t}^{\prime}$ in the first partition.
The remaining tokens, based on the principle of mutual exclusivity, are assigned to the red team $R_{t}^{\prime}$. The second partition continues to allocate words to the green list $G_{t}$ from the remaining vocabulary until the combined size of the green teams from both steps reaches the predefined limit, $\gamma$. The rest of the process follows the steps outlined in the vanilla watermarking of Alg. \ref{algo:encode}. 
\begin{algorithm}[h]
  \small
   \caption{WatME Encoding}
    \label{algo:WatME-encode}
    \begin{algorithmic} 
    \STATE \textbf{Input:}  prompt $x_{1}\cdots x_{m}$, 
    green list size $\gamma \in (0, 1)$, 
    watermark strength $\delta > 0$.

    \FOR{$t=0,1,\cdots, T-1$}
    \STATE 
    \begin{enumerate}[itemsep=0pt, topsep=0pt]
    \item Get the logit $\boldsymbol{\ell}_{t} \in \mathbb{R}^{|\mathcal{V}|}$ from $\cM$.
    \item Use seed from the last token, split each cluster   $C_i$ in parallel   into  green list $G_{it}'$ (of size $|C_i|\gamma$) and red list $R_{it}'$ (of size $(1-\gamma)|C_i|$) . 
Let $G_t' = \cup_i G_{it}'$ and $R_t' = \cup_i R_{it}'$. 
    \item Partition the remaining part $\mathcal{V} \setminus \mathcal{S}$ into a green list $G_t$ of size $\gamma |V| - |G_{t}^{\prime}|$ and a red list $R_t$ of size $(1-\gamma)|V| - |R_{t}^{\prime}|$. 
    \item Merge lists from the previous two steps: $G_t = G_t \cup G_{t}^{\prime}$ and $R_t = R_t \cup R_{t}^{\prime}$. 
    \item Add $\delta$ to the elements of logit $\boldsymbol{\ell}_t$  corresponding  to the green list, then softmax. \vspace*{-2.8mm} 
    $$\hat{\boldsymbol{p}}_t = softmax(\boldsymbol{\ell}_t[i] + \delta), i\in G_t$$
    \vspace*{-7.5mm}
    \item Sample the next token $y_{t+1}$ from $\hat{\boldsymbol{p}}_t$.  
    \end{enumerate}
    \ENDFOR
    \STATE {\bfseries Output:} watermarked text $y_{1} \cdots y_{T}$.
    \end{algorithmic}
\end{algorithm}

The WatME detection algorithm is unchanged, except the green list calculation now uses Alg. \ref{algo:encode}.

\subsection{Theoretical Analysis}
We provide a mathematical analysis demonstrating how WatME outperforms the conventional method, focusing on the 'green' team's expressiveness and the probability of high-quality sampling.

\begin{definition}[Semantic Entropy]
\label{def:Semantic_Richness}
Let $\mathcal{V}$ represent the vocabulary of a language model. We define the semantic entropy of $\mathcal{V}$, denoted by $H_{sem}(\mathcal{V})$, as the entropy of the semantic distribution across $\mathcal{V}$. This entropy quantifies the diversity and richness of meanings expressible by $\mathcal{V}$. Consequently, a higher value of $H_{sem}(\mathcal{V})$ signifies a vocabulary with greater semantic richness.
\end{definition}

\begin{definition}[Intrinsic Expressiveness]
\label{def:Intrinsic_Expressiveness}
It is assumed that a language model $\mathcal{M}$, with a vocabulary $\mathcal{V}$ characterized by high semantic entropy as indicated by $H_{sem}(\mathcal{V})$, possesses an enhanced intrinsic expressive capacity. This capacity is unaffected by the output distribution of $\mathcal{M}$ and is due to the extensive semantic capabilities of $\mathcal{V}$, which endow $\mathcal{M}$ with the potential for stronger expressive abilities.
\end{definition}

\begin{assumption}
\label{hypo:strongwatermark}
We consider practical scenarios that require high detectability, and thus a large value of $\delta$. In such a strong watermarking scenario, tokens from the green list are more probable to be used than those from the red list.
\end{assumption}

\paragraph{Note:}Assumption \ref{hypo:strongwatermark} establishes the foundational premise of text watermarking's effectiveness.

Building upon the Definitions and Assumption, we derive the following theorem.

\begin{theorem}
\label{thm:xmark-better}
Consider that $\boldsymbol{p}_{t} \in \mathbb{R}^{|\mathcal{V}|}$ represents the predicted distribution of the model $\mathcal{M}$ at decoding time $t$.
Let $w_i$ denote the token with the $i^{th}$ highest probability in $\boldsymbol{p}_{t}$. The higher the rank of a token (i.e., the smaller $i$ is), the more suitable it is to be selected. Under the conditions of Assumption \ref{hypo:strongwatermark}, the WatME watermarking method is more likely to select a suitable token compared to the vanilla watermarking method.
\end{theorem}

\begin{theorem}
\label{theorem:expressive_power_watme}
Given a fixed proportion $\gamma$ of the green team, the expressive power of a language model $\mathcal{M}$ employing the WatME exceeds that of one utilizing a vanilla watermarking approach.
\end{theorem}

These theorems highlight two advantages of WatME; their proofs are in the Appendix \ref{sec:proof}.

\section{Impact on Emergent Abilities} 
The majority of research on text watermarking utilizes the C4 dataset \cite{dodge2021documenting} as a basis for testing perplexity (PPL).
However, watermarking not only impacts the fluency of text generation but also holds the potential to influence LLMs on a broader scale, such as emergent abilities. These unique abilities intrinsic to LLMs garner significant interest from users and stimulate curiosity within the research community. However, they are often overlooked in the field of text watermarking.

Although a consensus definition is lacking, emergent abilities are typically characterized in many studies \cite{brown2020gpt3,wei2022emergent,yu2023generate} as a model's capacity to perform specific tasks without training. In light of this, we propose to test and compare the performance of WatME and Vanilla watermark algorithms on different tasks using prompting technologies.

To comprehensively test the impact of watermarking on these abilities, we attempt to categorize it into different scenarios for a more exhaustive examination. Specifically, we draw upon Cattell's cognitive theory \cite{cattell_theory_1963}, which bifurcates intelligence into crystallized and fluid intelligence. 
Crystallized intelligence corresponds to the model's utilization of learned knowledge and experience, while fluid intelligence involves logical thinking and solving problems. Correspondingly, we propose to examine crystallized intelligence through an assessment of the model's knowledge capabilities, and fluid intelligence through its ability to reason and solve mathematical problems.

\paragraph{Knowledge Capability.} 
To evaluate the model's mastery of world knowledge, we employ TruthfulQA \cite{lin2022truthfulqa}, a benchmark designed to test if LLMs can generate truthful and informative answers. 
We select the generation setting.
\paragraph{Reasoning Capability.} 
We employ the GSM8K dataset to assess the model's chain-of-thought reasoning. Comprising 8K arithmetic and math problems, it provides a platform for evaluating reasoning performance. Aligned with the CoT Hub prompt \cite{fu_chain}, our evaluations include few-shot scenarios that prompt the model to demonstrate reasoning and generate thought chains.

\section{Experiments}

\subsection{Experimental Setups}

\begin{table*}[!t]

\setlength{\tabcolsep}{1.3mm}
\small
\centering
\begin{tabular}{l|cc|cccc|cc}
\toprule
\multirow{2}{*}{\textbf{Model}} & \multicolumn{2}{c|}{\textbf{GSM8K}} & \multicolumn{4}{c|}{\textbf{TruthfulQA}} & \multicolumn{2}{c}{\textbf{C4}} \\
\cmidrule(lr){2-3} \cmidrule(lr){4-7} \cmidrule(lr){8-9}
 & \bf Acc. & \bf AUROC & \bf True. & \bf Info. & \bf True.*Info. & \bf AUROC & \bf PPL & \bf AUROC \\
\midrule
\textsc{Llama2-7b} & 11.22 & - & 95.10 & 92.78 & 88.23 & - & 4.77 & - \\

\addlinespace 
\quad \texttt{+} \textsc{KGW-Mark} & 5.61$_{-50.0\%}$ & 0.8886 & 57.16$_{-39.9\%}$ & 84.33$_{-9.1\%}$ & 48.20$_{-45.4\%}$ & 0.8416 & 7.00 & 0.9724 \\

\quad \texttt{+} \textsc{Gumbel-Mark} & 7.28$_{-35.1\%}$ & 0.9121 &  45.90$_{-51.7\%}$ & 92.78$_{-0.0\%}$ & 42.59$_{-51.7\%}$ & 0.4931 & 39.93 & 0.9422 \\

\quad \texttt{+} \textsc{Unbiased-Mark} & 10.24$_{-8.7 \%}$ & 0.5478 & 44.06$_{-53.7\%}$ & 93.76$_{+1.1\%}$ & 41.43$_{-53.0\%}$ & 0.5051 & 15.62 & 0.5451 \\

\quad \texttt{+} \textsc{Provable-Mark} & 5.16$_{-54.01\%}$ & 0.9052 & 64.14$_{-32.6\%}$ & 91.68$_{-1.2\%}$ & 58.80$_{-33.4\%}$ & 0.9555 & 10.21 & 0.9623 \\

\addlinespace 
\quad \texttt{+} \textsc{WatME}$_{dictionary}$ & 9.17$_{-18.3\%}$ & 0.8995 & 69.28$_{-27.2\%}$ & 88.25$_{-4.9\%}$ & 61.14$_{-30.7\%}$ & 0.8848 & 5.32 & 0.9804 \\

\quad \texttt{+} \textsc{WatME}$_{prompting}$ & 5.84$_{-48.0\%}$ & 0.9128 & 55.83$_{-41.3\%}$ & 95.10$_{+2.5\%}$ & 50.39$_{-42.9\%}$ & 0.8659 & 6.89 & 0.9724 \\

\midrule

\multicolumn{7}{c}{} \\[-0.5em] 

\textsc{Vicuna-v1.5-7B} & 17.51 & - & 93.88 & 87.27 & 81.92 & - & 10.77 & - \\

\addlinespace 
\quad \texttt{+} \textsc{KGW-Mark} & 13.87$_{-20.8\%}$ & 0.7870 & 74.05$_{-21.1\%}$ & 87.52$_{+0.3\%}$ & 64.81$_{-20.1\%}$ & 0.7417 & 11.62 & 0.9679 \\

\quad \texttt{+} \textsc{Gumbel-Mark} & 9.02$_{-48.5\%}$ & 0.7077 & 68.30$_{-27.2\%}$ & 87.27$_{-0.0\%}$ & 59.61$_{-27.2\%}$ & 0.4647 & 48.93 & 0.8617 \\

\quad \texttt{+} \textsc{Unbiased-Mark} & 17.89$_{+2.2\%}$ & 0.5508 & 70.38$_{-25.0\%}$ & 88.86$_{+1.8\%}$ & 62.54$_{-23.7\%}$ & 0.4855 & 19.93 & 0.5000 \\

\quad \texttt{+} \textsc{Provable-Mark} & 12.21$_{-30.27\%}$ & 0.8020 & 74.42$_{-20.7\%}$ & 96.70$_{+10.8\%}$ & 71.96$_{-12.2\%}$ & 0.8796 & 10.21 & 0.9582 \\

\addlinespace 
\quad \texttt{+} \textsc{WatME}$_{dictionary}$ & 14.78$_{-15.6\%}$ & 0.8044 & 78.95$_{-15.9\%}$ & 97.43$_{+11.6\%}$ & 76.92$_{-6.1\%}$ & 0.7897 & 10.96 & 0.9582 \\

\quad \texttt{+} \textsc{WatME}$_{prompting}$ & 16.22$_{-7.4\%}$ & 0.7843 & 69.65$_{-25.8\%}$ & 97.45$_{-11.5\%}$ & 67.87$_{-17.2\%}$ & 0.7396 & 11.54 & 0.9519 \\
\bottomrule
\end{tabular}
\caption{Performance comparison of Llama2-7B and Vicuna-v1.5-7B under different watermarking algorithms. 
}
\label{tab:all_results}
\vspace{-0.25cm}
\end{table*}

\paragraph{Evaluation Metrics} 
To evaluate detection performance, following previous work, we use the Area Under the Receiver Operating Characteristic curve (\textit{AUROC}), a well-established metric for binary classifiers. For mathematical reasoning tasks, we use \textit{Accuracy} to assess the correctness of the model's solutions. In our evaluation of the TruthfulQA dataset, following \citet{lin2022truthfulqa}, we use the trained GPT-Truth and GPT-Info scorers, assessing the model's capacity to generate both truthful and informative responses. 
Given the potential trade-off between these two perspectives, the product of Truth and Information (\textit{Truth.*Info.}) is commonly used as an overall measure of performance.
On the C4 dataset, we report Perplexity (PPL).
\vspace{-0.2cm}

\paragraph{Baselines}
We compared our model with four established baselines. First, KGW-Mark (Vanilla watermarking) \cite{watermark}, which categorizes teams into 'red' and 'green' to facilitate detection. Second, Gumbel-Mark \cite{stanford2023robust}, which employs a Gumbel-Softmax distribution to introduce stochasticity into the watermarking process. Third, Unbiased-Mark \cite{hu2024unbiased}, which implements reweighting techniques to maintain the expected output distribution of the LLM during watermarking. Lastly, Provable-Mark \cite{zhao2023provable}, which uses a fixed hash key during watermarking to achieve provably better performance.\vspace{-0.2cm}

\paragraph{Models}
We utilized two distinct types of LLMs for experimentation: the non-aligned Llama2 model \cite{llama2}, and the aligned Vicuna v1.5 model \cite{vicuna}. The majority of the results reported in this paper were obtained using the 7B version of the models.

Further setup details are in Appendix \ref{sec:setup}.

\subsection{Main Results}
\paragraph{Greater Impact on Emergent Abilities than Fluency}
The experimental evidence suggests that watermarking notably hinders the emergent abilities of LLMs much more than fluency (see Table \ref{tab:all_results}).
Specifically, the non-aligned Llama2 model experienced a decline in performance exceeding 50\% on both the GSM8K and TruthfulQA benchmarks. 
In contrast, the aligned model, Vicuna, demonstrated relative resilience but still incurred performance reductions greater than 20\% on these benchmarks. This demonstrates the adverse impact of Vanilla Watermarking on the knowledge and reasoning capabilities of LLMs, with reasoning showing a marginally greater susceptibility. 
\vspace{-2.0mm}

\paragraph{Superiority of WatME over baselines in Preserving Emergent Abilities}
Across all models and benchmarks, the WatME consistently outperformed baseline watermarking methods. For the Llama2 model, WatME mitigated performance degradation by 16.8\% on GSM8K and by 14.7\% on TruthfulQA compared to the strongest baseline. Similarly, for the Vicuna model, the reductions were 13.4\% and 14.0\%, respectively. These outcomes underscore WatME's significant effectiveness in preserving the emergent capabilities of LLMs without compromising performance as significantly as other methods. 
\vspace{-2.0mm}

\paragraph{Comparable Detection Performance of WatME}
Despite the trade-off between text quality and detection performance, WatME's detection efficacy matched that of the Vanilla Watermark while also enhancing model capabilities, as evidenced by similar AUROC scores—suggesting our algorithm attained a better equilibrium than the baseline. In contrast, the Gumbel-Mark method noticeably compromised detection performance, particularly in aligned models and when generating short responses (TruthfulQA). 
Additionally, more performance results under different watermark strengths are presented in Discussion \ref{sec:Hyperparameters}.
\vspace{-2.0mm}

\paragraph{Distinct Advantages of WatME Variations}
It is evident that different WatME variations exhibit unique strengths; 
The 'dictionary' variant outperformed in the \textit{Accuracy} and \textit{Truthfulness} scores, while the 'prompting' variant excelled in the \textit{Informativeness}. The integration of these variants may offer a fruitful avenue for future research. 
For a comprehensive understanding, a manual analysis of lexical clusters derived from these methods is presented in the Discussion \ref{sec:cluster_quality}.
\vspace{-2.0mm}

\paragraph{Alignment Diminishes Watermark Effectiveness}
Surprisingly, aligned models showed significantly greater resistance to watermarking effects than non-aligned models.
Specifically, Vicuna 1.5's performance dipped 30\% less than Llama2's across all benchmarks, coupled with a 10\% lower watermark detection performance.
To understand the underlying reasons for these differences, we analyzed the output distribution discrepancies between aligned and unaligned models in the Discussion \ref{sec:distribution_ana}.

\section{Discussion}
\vspace{-0.05cm}

\begin{figure}[t]
    \centering
    \includegraphics[width=\linewidth]{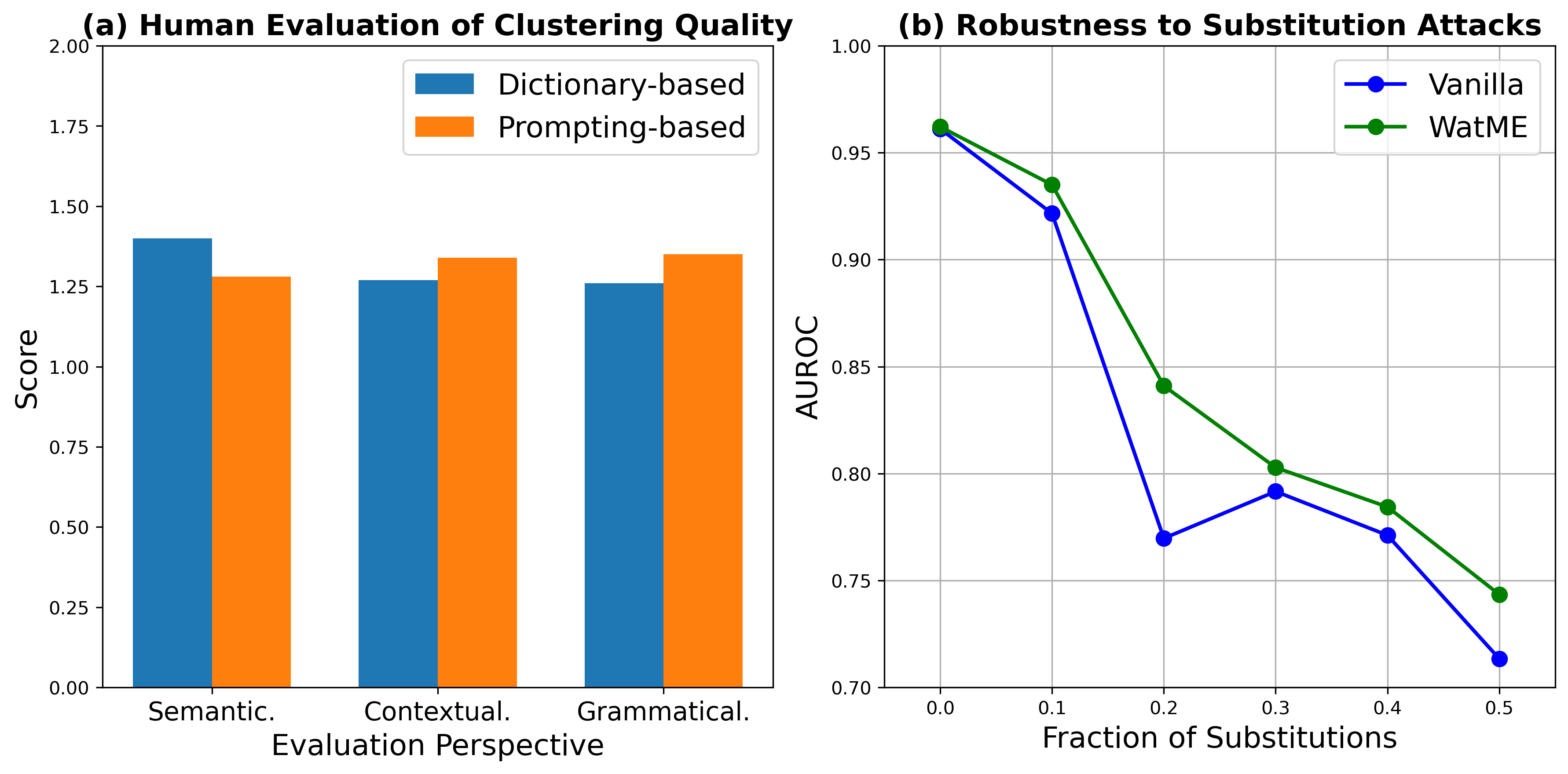} 
    \caption{
    (a) Human evaluation for the quality of clusters mined by varied methods and (b) testing detection robustness against substitution attacks.}
    \label{fig:human_robust}
    \vspace{-0.40cm}
\end{figure}

\subsection{Analysis of Clustering Methods}
\label{sec:cluster_quality}

To analyse redundant clusters from diverse methods, we set evaluation criteria to ensure analytical rigour.
These criteria spanned \textit{semantic consistency}, \textit{contextual appropriateness}, and \textit{grammatical consistency}, which are essential aspects of cluster quality. 
 Two annotators used a rating scale of {0, 1, 2} to annotate the clusters.
A score of '2' indicated high or ideal consistency, '1' denoted moderate or usable consistency, and '0' identified low or unusable consistency within a cluster. The kappa value for the annotations is 0.657.
Figure \ref{fig:human_robust}(a) shows both methods met usability, but fell short of ideal effectiveness. 
The dictionary approach was superior in semantic coherence due to its utilization of lexical databases. 
Conversely, the prompting method outperformed in contextual and grammatical consistency, reflecting the varied linguistic corpus training of LLMs. 
This suggests the potential benefits of a combined approach, a topic reserved for future research.

\subsection{Robustness Against Attacks}
In addition to affecting the performance of LLMs, watermarks are also vulnerable to attacks aimed at their removal. To evaluate the robustness of our method, we conducted tests against two prevalent types of attacks: substitution attacks and paraphrase attacks.
For the substitution attack, we evaluated 200 examples from GSM8k, with various token replacement ratios. As shown in Figure \ref{fig:human_robust}(b), WatME consistently outperformed the baseline method in the robustness of detection across different levels of token replacement.
For paraphrase attacks, we use a powerful paraphraser, llama-2-chat-13B, to extensively rewrite the watermarked text generated by llama-2-7b. We provided it with the prompt: "Please paraphrase the following text, altering the wording significantly yet preserving the original meaning and length." We then subjected our system to these rewritten samples using 200 entries from both GSM8k and TruthfulQA. The results are presented in Tables \ref{tab:combined_results}.

We offer two perspectives to understand the robustness of WatME: (1) Intuitively, for substitution attacks, the effect on watermarking depends on whether it triggers a token swap between the 'red' and 'green' teams: a swap affects the detection, while no swap means the watermark remains intact. With KGW-Mark, semantically similar tokens may be allocated to one team, resulting in a substitution invariably causing a swap. In contrast, WatME is intentionally designed to prevent this scenario. Therefore, the likelihood of a red-green swap—and consequently the impact on the watermark—is reduced in WatME compared to KGW. (2) From an encryption viewpoint, whereas KGW-Mark relies on a single division between teams, WatME employs multiple divisions—the number of clusters plus one (|C|+1), as outlined in Algorithm \ref{algo:encode}. Though these multiple partitions are computationally equivalent to a single partition due to efficient parallel matrix operations (explained in Appendix \ref{sec:time_ana}), they introduce a higher level of complexity and robustness to the encryption process.

\begin{table}[t!]
\setlength{\abovecaptionskip}{5pt}   
\setlength{\belowcaptionskip}{0pt}
\setlength{\tabcolsep}{0.52mm}
\centering

\begin{tabular}{@{}lccc@{}}
\toprule
Method & Dataset & Original & Para. Attack \\ \midrule
KGW-Mark & GSM8k & 0.885 & 0.745 \\
WatME &  & 0.955 & 0.910 \\
\midrule

KGW-Mark & TruthfulQA & 0.924 & 0.528 \\
WatME &  & 0.949 & 0.673 \\ \bottomrule
\end{tabular}
\caption{Detection robustness against paraphrasing attacks.}
\label{tab:combined_results}
\vspace{-0.3cm}
\end{table}

\subsection{Performance Trade-offs at different Delta} 
\label{sec:Hyperparameters}
\begin{figure}[t]
    \centering
    \includegraphics[width=\linewidth]{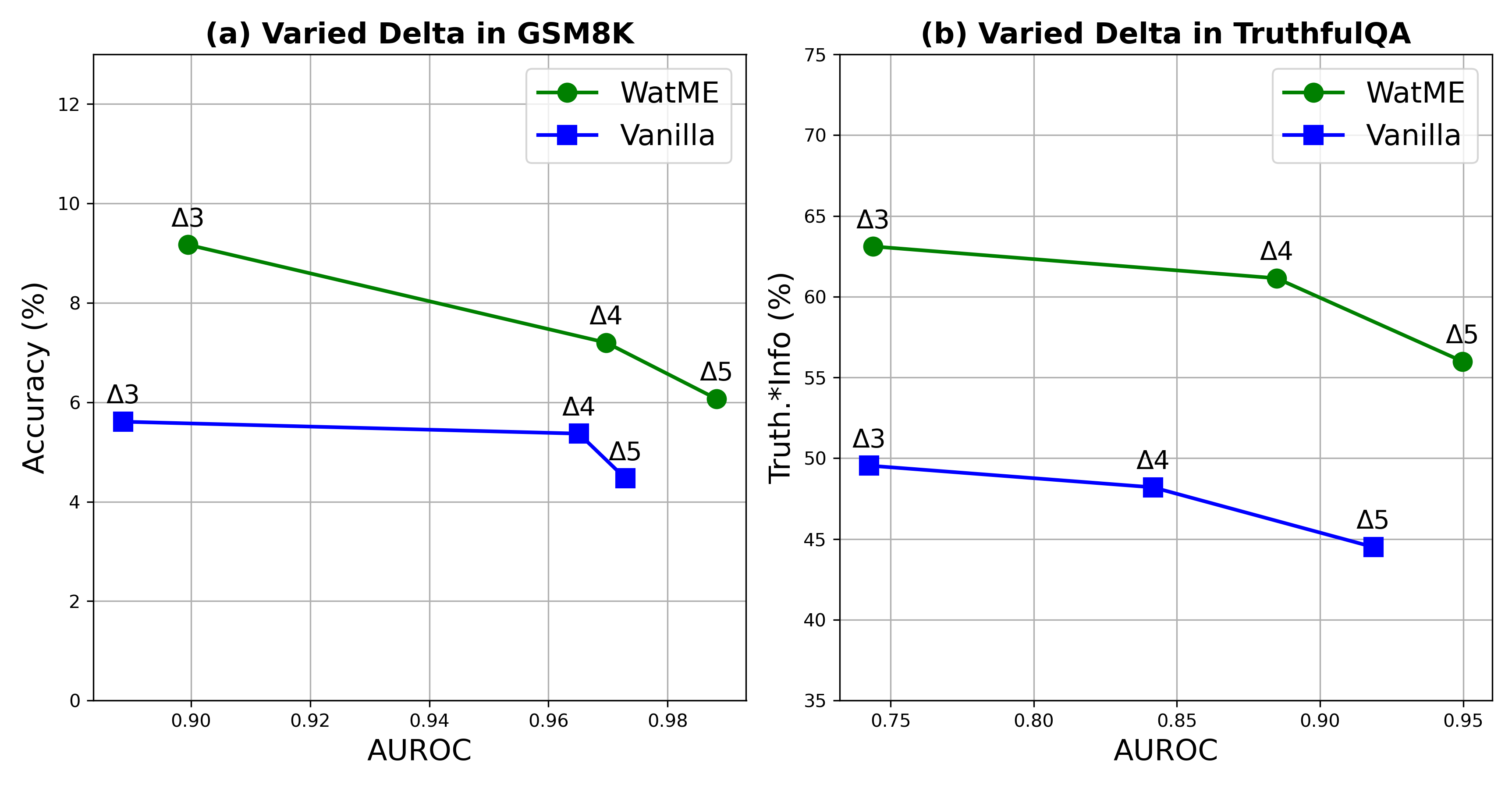} 
    \caption{Performance trade-offs comparison between WatME and Vanilla Watermark on TruthfulQA and GSM8K at different Delta ($\Delta$) values. }
    \label{fig:hyperparam}
  \vspace{-0.30cm}
\end{figure}

The efficacy of the Watermark is influenced by the hyperparameter, Delta, which controls the watermark strength. 
An increase in Delta facilitates easier watermark detection but at the cost of severer impact on the LLMs.
We analyse the TruthfulQA and GSM8K datasets. Figure \ref{fig:hyperparam} shows WatME consistently achieved a more favourable balance between watermark robustness and LLM performance across various Delta settings, surpassing Vanilla Watermark. Notably, the performance curves of WatME are strictly better than that of Vanilla, indicating that at equivalent watermark strengths, WatME always maintains superior performance compared to Vanilla Watermark. 

\subsection{Aligned vs Unaligned Models}
\label{sec:distribution_ana}
\begin{figure}[t]
    \centering
    \includegraphics[width=\linewidth]{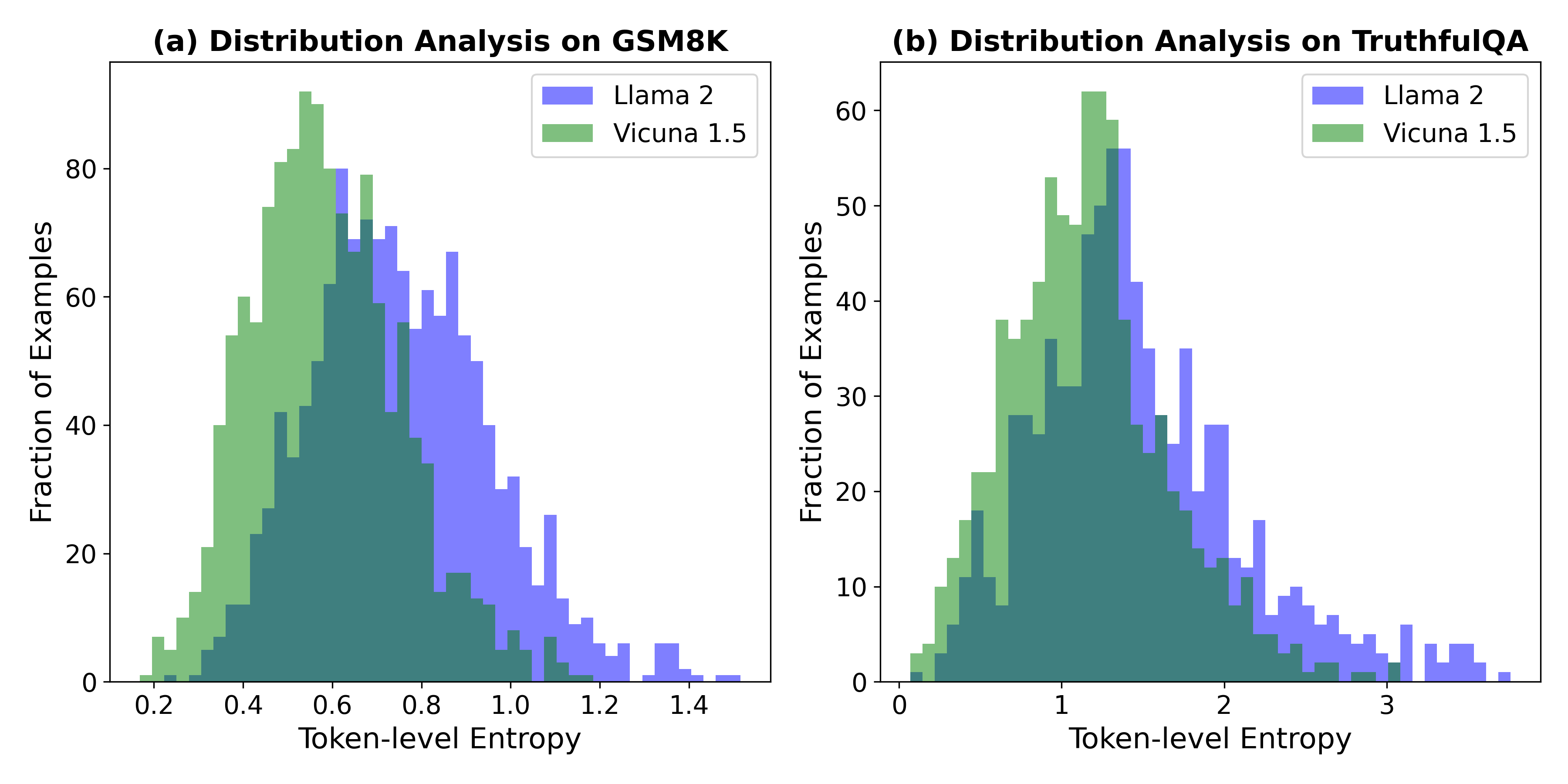} 
    \caption{Token-level entropy distributions for aligned (green) and unaligned (blue) models on GSM8K and TruthfulQA benchmarks.}
    \label{fig:dist_ana}
    \vspace{-0.30cm}
\end{figure}

Our examination of the response sensitivity to watermarking in aligned and unaligned models involved analyzing their output distributions on the TruthfulQA and GSM8K datasets.
We computed the average entropy for tokens in the generated answers and found that aligned models exhibit markedly lower entropy, suggesting more deterministic response patterns, as illustrated in Figure \ref{fig:dist_ana}. This pronounced certainty in aligned models' outputs presents a challenge for watermarking because of the limited variability that is essential for effective watermark encoding.

\section{Conclusion}
This study explores the impact of watermarking on the emergent abilities of LLMs—an aspect often neglected in the field. Our findings highlight the considerable adverse effects of traditional watermarking methods on LLMs' emergent abilities, including knowledge recall and logical reasoning.

In response, we introduced WatME—a novel watermarking approach that leverages lexical redundancy. Theoretical analysis and comprehensive empirical results indicate WatME consistently preserves the expressive power of LLMs without compromising detection performance, enabling developers to encode watermarks with less disruption to user experience.

The advancements with WatME mark a stride in lossless watermarking, enabling developers to encode watermarks with less disruption to user experience.
We hope to promote a better equilibrium between regulatory compliance and user satisfaction in LLM development.

\section*{Limitations}
In this section, we discuss the limitations of this work from two perspectives.

Firstly, although  WatME represents a step toward lossless watermarking, it is not entirely loss-free. The introduction of a controlled bias, inherent to watermarking methods, subtly alters the generated data. This compromise is a critical consequence as it diverges from the ideal of a completely lossless system. 
This deviation poses a dilemma for developers weighing the benefits of watermarking against potential user experience and regulatory trade-offs. Future work aims to bridge this gap, enhancing the WatME method to maintain output integrity and broaden its appeal for practical implementation.

Secondly, while our method is designed to be language-agnostic, the empirical validation presented in this work is limited to models processing the English language. 
We acknowledge that the applicability of watermarking across various linguistic contexts is critically important. Future investigations will endeavour to broaden the scope to include more languages, ensuring the generalizability and effectiveness of our approach in a multilingual context.

Thirdly, the challenge of watermarking in low-entropy scenarios remains an open problem. Our dataset encompasses a range of scenarios, including low-entropy situations where outcomes are more predictable and our methodology remains effective. However, embedding watermarks in text with universally recognized, low-entropy answers poses significant challenges, highlighting the need for further investigation into constructing and testing methodologies for low-entropy corpora.

Lastly, our LLMs-based cluster generation approach is influenced by the robustness of the prompting methods. Different prompt constructions can lead to varying outcomes \cite{pmlr-v139-zhao21c,chen-etal-2023-towards-robust,chen2024simple}, represents a limitation that warrants further discussion and exploration in future work.

Despite these limitations, we believe our work serves as a significant catalyst for the field, contributing positively to the advancement of more lossless and detectable text watermarking techniques.

\bibliography{anthology,custom}
\bibliographystyle{acl_natbib}

\appendix

\section*{Appendix}
\label{sec:appendix}
\addcontentsline{toc}{section}{Appendix}

\section{Algorithms of Watermark}
\label{sec:algorithms}
This section presents detailed algorithms for the watermark encoding and detection processes as outlined in \cite{watermark}. Algorithm \ref{algo:encode} delineates the procedure for encoding a watermark into the output sequence generated by a language model. Conversely, Algorithm \ref{algo:detect} explicates the method for detecting and confirming the watermark's presence within generated sequences.

\begin{algorithm}[h]
    \caption{Vanilla Watermark Encoding}
    \label{algo:encode}
    \begin{algorithmic} 
    \STATE \textbf{Input:}  prompt $x_{1}\cdots x_{m}$, \\

    \hspace{10.5mm} green list size $\gamma \in (0, 1)$, 

    \hspace{10.5mm} watermark strength $\delta > 0$.

    \FOR{$t=0,1,\cdots, T-1$}
    \STATE 
    \begin{enumerate}[itemsep=0pt, topsep=0pt]
    \item Get the logit $\boldsymbol{\ell}_{t} \in \mathbb{R}^{|\mathcal{V}|}$ from $\cM$.
    \item Use the hash of the previous token as the random seed 
    to partition the vocabulary of $\cM$ into a ``green list'' $G_t$ of size $\gamma |\mathcal{V}|,$ and a ``red list'' $R_t$ of size $(1-\gamma)|\mathcal{V}|$. 
    \item Add $\delta$ to each green list logit and then apply softmax to the modified logits. \vspace*{-2.8mm} 
    $$\hat{\boldsymbol{\ell}}_t[i] := \boldsymbol{\ell}_t[i] + \delta, i\in G_t$$ \vspace*{-7mm}
    $$\hat{\boldsymbol{p}}_t = softmax(\hat{\boldsymbol{\ell}}_t)$$ 
    \vspace*{-7.5mm}
    \item Sample a next token $y_{t+1}$ from $\hat{\boldsymbol{p}}_t$.  
    \end{enumerate}
    \ENDFOR
    \STATE {\bfseries Output:} watermarked text $y_{1} \cdots y_{T}$.
    \end{algorithmic}
\end{algorithm}

\begin{algorithm}[h]
   \caption{Vanilla Watermark Detection}
   \label{algo:detect}
\begin{algorithmic}
   \STATE {\bfseries Input:} text $\boldsymbol{y}$, detection threshold $\tau$. \\
   \STATE 1.\, Use the previous token to find the ``green list'' $G_t$ at the step $t$ as in Alg. \ref{algo:encode}. 
   \STATE 2.\, Calculate the number of green tokens in $\boldsymbol{y}$ as $|\boldsymbol{y}|_G = \sum_{t=1}^n \mathbbm{1}(y_t \in G)$.
   \STATE 3.\, Compute the $z$-statistic: \vspace*{-2.8mm} 
   $$z_{\boldsymbol{y}}=\left(|\boldsymbol{y}|_G-\gamma |\mathcal{V}|\right) / \sqrt{|\mathcal{V}| \gamma(1-\gamma)}.$$ \vspace*{-6.8mm}
   \STATE 4.\, {\bfseries if} $z_{\boldsymbol{y}} > \tau$ {\bfseries then return}  1 (watermarked).
   \STATE 5.\, {\bfseries else return} 0 (unwatermarked).
   \STATE {\bfseries Output:} 0 or 1
\end{algorithmic}
\end{algorithm}

\section{Prompt for Cluster Mining}
\label{cluster_mining_prompt}
\begin{figure*}[t]
    \centering
    \includegraphics[width=\linewidth]{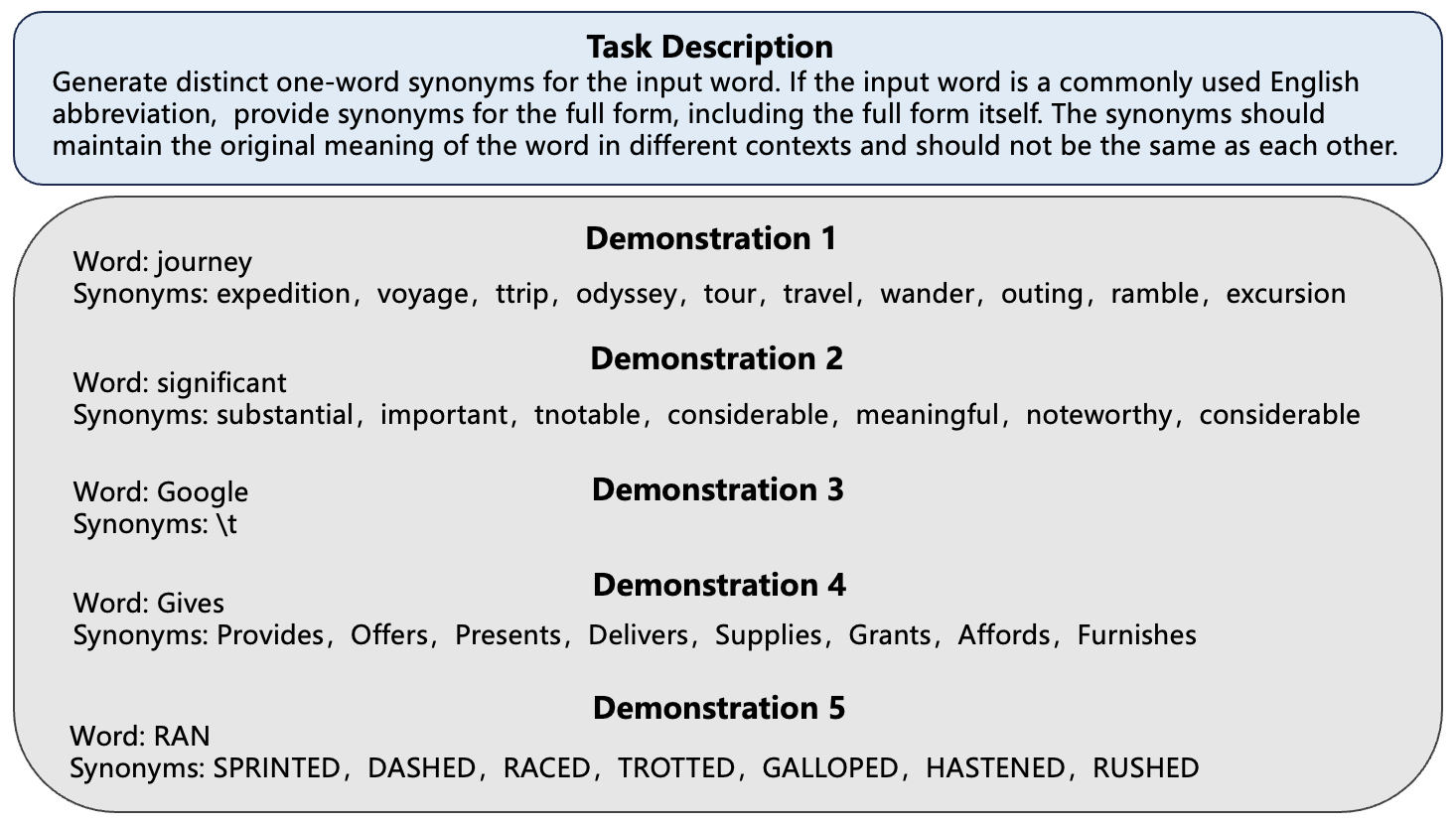} 
    \caption{Few-Shot Demonstration of Synonym Generation using LLMs.}
    \label{fig:mine_prompt}
\end{figure*}

To facilitate the generation of synonym clusters, we employed Llama2-13B-chat. The approach involved crafting a prompt (Figure \ref{fig:mine_prompt}) that combines a clear task description with a set of demonstrations designed to illustrate the desired task. By presenting the model with a few-shot example, we primed Llama2-13B-chat to understand and perform the specific task of synonym generation. The few-shot prompt was crucial for the model to recognize the pattern and replicate it for new target words, thus enabling the mining of synonym clusters effectively.

\section{Proofs of Theorems}
\label{sec:proof}
In this section, we present the detailed proofs of the theorems introduced before. Each theorem is treated in its respective subsection.

\subsection{Proof of Theorem \ref{thm:xmark-better}}
\begin{proof}
We begin the proof by considering \( i = 1, 2 \).

\noindent \textbf{Case I: where $w_1$ is in the green list ($G_t$):}

If $w_1 \in G_t$, then both watermarking methods are lossless because they can select the most suitable token $w_1$. 

\noindent \textbf{Case II: where $w_1$ is in the red list ($R_t$):}

We consider $w_2$, which may or may not be a synonym of $w_1$:

\textbf{Sub-case i:} $w_2$ is not a synonym of $w_1$.

If $w_1 \notin G_t$ and $\centernot\exists C_i \in \mathcal{C}$ s.t. $w_1, w_2 \in C_i$, then according to Algo. \ref{algo:WatME-encode} we have: \vspace{-0.14cm}
\begin{align*} 
P_{WatME}(w_2 \in G_t) &= P_{watermark}(w_2 \in G_t). 
\end{align*}
In this case, the two methods are the same.

\textbf{Sub-case ii:} $w_2$ is a synonym of $w_1$.

If $w_1 \notin G_t$ and $\exists C_i \in \mathcal{C}$ s.t. $w_1, w_2 \in C_i$, then according to Algo. \ref{algo:WatME-encode} we have: \vspace{-0.14cm}
\begin{align*} 
P_{WatME}(w_2 \in G_t) &> P_{watermark}(w_2 \in G_t).
\end{align*}
Based on Assumption \ref{hypo:strongwatermark}, WatME is more likely to select the suitable token.
Combining these cases, the theorem is proven. It should be noted that while this proof explicitly considers the cases for $i=1,2$, the logic extends to any arbitrary value of $i$.
\end{proof}

\subsection{Proof of Theorem \ref{theorem:expressive_power_watme}}
\begin{proof}
Let us define the vocabulary \( V \) with synonym clusters \( S = \{C_1, \ldots, C_n\} \), where \( \bar{C} \) represents the set of non-synonymous, unique words. According to Algs \ref{algo:encode} and \ref{algo:WatME-encode}, WatME maintains a constant number of distinct semantic representations, quantified as \( n + \gamma \cdot \lvert \bar{C} \rvert \). In contrast, the semantic token count of standard watermarking algorithms is lower than this figure. According to Definition \ref{def:Semantic_Richness} the disparity in semantic entropy between the two methodologies is thus evident. Given Definition \ref{def:Intrinsic_Expressiveness}, the increased semantic entropy inherent to WatME confirms the theorem.
\end{proof}

\section{Time Complexity Analysis} \label{sec:time_ana}
The conventional algorithm necessitates a partition of the vocabulary per decoding operation, which results in a time complexity of $O(|V|)$. Our method incorporates two partitioning stages: initially targeting the redundant cluster, followed by the remaining vocabulary. During the first stage, we pad the cluster into a 2D matrix and conduct parallel sampling. The subsequent stage aligns with the procedures of the Vanilla algorithm. Consequently, the time complexity of our method remains at $O(|V|)$.

\section{Setup Details} 
\label{sec:setup}
In our experiments, we used prompts from the CoT hub \cite{fu_chain} for the GSM8K dataset and the original prompts from TruthfulQA \cite{lin2022truthfulqa}. The Llama2 model was evaluated using its original prompt format to maintain consistency. Greedy decoding was employed as the strategy for all tasks, with maximum decoding lengths set at 128 tokens for GSM8K and 50 tokens for TruthfulQA, which allowed for the complete generation of answers within the datasets.

To account for the differing answer lengths in the GSM8K and TruthfulQA datasets, we fine-tuned the watermark hyperparameters. For GSM8K, with its longer answers aiding detection, we used a milder watermark intensity, setting gamma at 0.3 and delta at 3.0. Conversely, the brevity of answers in TruthfulQA complicates detection, necessitating a stronger watermark intensity—again with gamma at 0.3, but with delta increased to 4.0 to achieve satisfactory detection performance (AUROC > 0.7).

Evaluation metrics were carefully chosen: AUROC was calculated using the `sklearn` library, and for the assessment of GPT-Truth and GPT-Info, we utilized a fine-tuned Llama2-13B-chat model that demonstrated an accuracy above 93\% on the validation set. All model implementations were executed using the `transformers` library.

The hardware employed for these experiments consisted of a 40GB A100 GPU and a 32GB V100 GPU, ensuring sufficient computational power for model training and evaluation.

\section{Examples of Redundant Clusters} 
We present some examples of mined clusters at 
\ref{tab:redundant_clusters}.

\begin{table*}[htbp]
\centering
\begin{tabular}{@{}ll@{}}
\toprule
\textbf{Dictionary-based Method} & \textbf{LLM-based Method} \\ \midrule
'should', 'must', 'would' & 'must', 'ought', 'should' \\
'job', 'pursuit', 'operation', 'profession', 'career', 'employment', 'behavior' & 'job', 'task', 'work' \\
'inside', 'in' & '\_inside', '\_inner', '\_within' \\
\bottomrule
\end{tabular}
\caption{Examples of Redundant Clusters.}
\label{tab:redundant_clusters}

\end{table*}

\begin{table}[htbp]
\centering
\begin{tabular}{@{}lcc@{}}
\toprule
Method & ROUGE-L & AUROC \\ \midrule
ChatGLM 3-6b & 11.29 & - \\
+KGW-Mark & 8.89 & 0.8415 \\
+WatME\_prompting & 10.23 & 0.8514 \\
\bottomrule
\end{tabular}
\caption{CLTS Results}
\label{tab:clts_results}
\end{table}

\section{Multilingual Performance Testing}
We expand our evaluation to include the Chinese Long Text Summarization Dataset (CLTS) and a bilingual Large Language Model (LLM), ChatGLM3-6b. This model employs Byte Pair Encoding (BPE) tokenization with a vocabulary size of 65k—double that of the Llama 2 model which has a 32k vocabulary size. Synonym mining, a critical step in our process, was conducted using the ChatGLM3-13B model. The performance of different watermarking methods was evaluated using the ROUGE-L and AUROC metrics, as shown in Table \ref{tab:clts_results}.
The results highlight that watermarking with WatME considerably enhances detection robustness compared to the baseline method, maintaining effectiveness despite varying levels of token replacement. This improvement underscores WatME's capability to integrate seamlessly without compromising the natural language generation quality.

\end{document}